\documentclass[10pt,twocolumn,letterpaper]{article}

\usepackage{cvpr}
\usepackage{times}
\usepackage{epsfig}
\usepackage{graphicx}
\usepackage{amsmath}
\usepackage{amssymb}
\usepackage{mathtools}
\usepackage{multirow}
\usepackage{colortbl}
\usepackage{xfrac}

\newcommand{\neuralnet}{\Phi}

\newcommand{\ainit}{\alpha_{\mathit{init}}}
\newcommand{\afin}{\alpha_{\mathit{fin}}}
\newcommand{\avis}{\alpha_{\mathit{vis}}}
\newcommand{\rvis}{r_{\mathit{vis}}}
\newcommand{\rinit}{r_{\mathit{init}}}
\newcommand{\rfin}{r_{\mathit{fin}}}
\newcommand{\rgt}{r_{\mathit{gt}}}
\newcommand{\cinit}{c_{\mathit{init}}}
\newcommand{\cfin}{c_{\mathit{fin}}}
\newcommand{\cvis}{c_{\mathit{vis}}}

\newcommand{\Deltabx}{\Delta_\mathbf{x}}
\newcommand{\Deltad}{\Delta_d}
\newcommand{\dmax}{d_{\mathit{max}}}
\newcommand{\mypm}{\mathbin{\smash{%
\raisebox{0.35ex}{%
            $\underset{\raisebox{0.5ex}{$\smash -$}}{\smash+}$%
            }%
        }%
    }%
}

\definecolor{Yellow}{rgb}{1,1, 0.7}

\usepackage[breaklinks=true,bookmarks=false]{hyperref}

\cvprfinalcopy 

\def\cvprPaperID{2957} 

\ifcvprfinal\fi
\begin{document}

\title{Pushing the Boundaries of View Extrapolation with Multiplane Images}
\author{Pratul P. Srinivasan$^1$
\qquad
Richard Tucker$^2$
\qquad
Jonathan T. Barron$^2$ \\
Ravi Ramamoorthi$^3$
\qquad
Ren Ng$^1$
\qquad
Noah Snavely$^2$ \\
\small{$^1$UC Berkeley, $^2$Google Research, $^3$UC San Diego}
}

\maketitle
\thispagestyle{empty}

\begin{abstract}
We explore the problem of view synthesis from a narrow baseline pair of images, and focus on generating high-quality view extrapolations with plausible disocclusions. Our method builds upon prior work in predicting a multiplane image (MPI), which represents scene content as a set of RGB$\alpha$ planes within a reference view frustum and renders novel views by projecting this content into the target viewpoints. We present a theoretical analysis showing how the range of views that can be rendered from an MPI increases linearly with the MPI disparity sampling frequency, as well as a novel MPI prediction procedure that theoretically enables view extrapolations of up to $4\times$ the lateral viewpoint movement allowed by prior work. Our method ameliorates two specific issues that limit the range of views renderable by prior methods: 1) We expand the range of novel views that can be rendered without depth discretization artifacts by using a 3D convolutional network architecture along with a randomized-resolution training procedure to allow our model to predict MPIs with increased disparity sampling frequency. 2) We reduce the repeated texture artifacts seen in disocclusions by enforcing a constraint that the appearance of hidden content at any depth must be drawn from visible content at or behind that depth.
\end{abstract}

\begin{figure}[!]
\begin{center}
\newcommand{\width}{1.0\linewidth}
\includegraphics[width=\width]{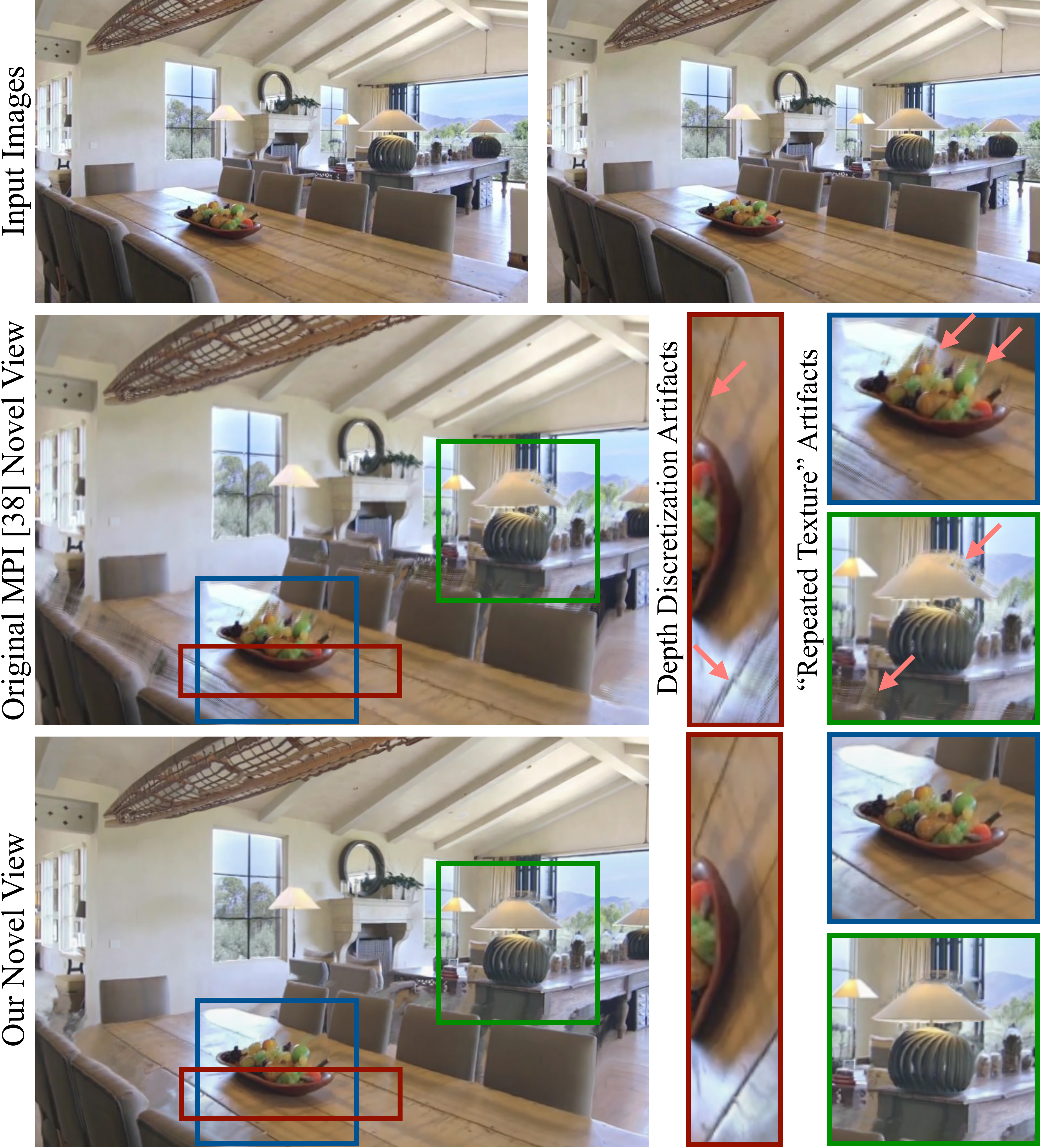} 
\caption{Given two input images taken from nearby viewpoints, our algorithm predicts an MPI scene representation that can render view extrapolations with disocclusions. Our model improves upon prior work in two specific ways: 1) We reduce depth discretization artifacts due to insufficient depth sampling, as seen in the red zoom of the wood table. 2) We mitigate the repeated texture artifacts produced by prior methods by predicting plausible hidden scene content, as shown in the blue and green zooms where we predict realistic textures behind the fruit bowl and lamp.}
\label{fig:teaser}
\end{center}
\vspace{-0.1in}
\end{figure}

\section{Introduction}

View synthesis, the problem of predicting novel views of a scene from a set of captured images, is a central problem in computer vision and graphics. The ability to render nearby views from a single image or a stereo pair can enable compelling photography effects such as 3D parallax and synthetic defocus blur. Furthermore, given a collection of images of a scene taken from different viewpoints, view synthesis could enable free-viewpoint navigation for virtual and augmented reality. 

However, there is still a long way to go. State-of-the-art view synthesis algorithms use their input images to estimate a 3D scene representation, which can then be reprojected to render novel views. This approach works well for content visible in the input images, but the quality of novel views degrades rapidly as the target viewpoint moves further away from the input views, thereby revealing more previously-occluded scene content. In this work, we study the problem of view extrapolation where regions of the rendered images observe disoccluded content, and focus specifically on demonstrating view synthesis from a stereo input.

We build upon a state-of-the-art deep learning approach for view synthesis~\cite{zhou18} that predicts a scene representation called a multiplane image (MPI) from an input narrow-baseline stereo pair. An MPI consists of a set of fronto-parallel RGB$\alpha$ planes sampled within a reference view camera frustum, as illustrated by Figure~\ref{fig:representation}. Diffuse volumetric scene representations such as the MPI are becoming increasingly popular for view synthesis for a number of reasons: 1) They can represent geometric uncertainty in ambiguous regions as a distribution over depths, thereby trading perceptually-distracting artifacts in those ambiguous regions for a more visually-pleasing blur~\cite{levoy88,penner17}. 2) They are able to convincingly simulate non-Lambertian effects such as specularities~\cite{zhou18}. 3) They are straightforward to represent as the output of a CNN and they allow for differentiable rendering, which enables us to train networks for MPI prediction using only triplets of frames from videos for input and supervision~\cite{zhou18}. 
In this work, we extend the MPI prediction framework to enable rendering high-quality novel views up to $4\times$ further from the reference view than was possible in prior work. Our specific contributions are:

\medskip
\noindent \textbf{Theoretical analysis of MPI limits (Section~\ref{sec:theory}).} We present a theoretical framework, inspired by Fourier theory of volumetric rendering and light fields, to analyze the limits of views that can be rendered from diffuse volumetric representations such as the MPI. We show that the extent of renderable views is limited by the MPI's disparity sampling frequency, even for content visible in both the input and rendered views, and that this ``renderable range'' increases linearly with the MPI's disparity sampling frequency. 

\medskip
\noindent \textbf{Improved view extrapolation for visible content (Section~\ref{sec:arch}).} View extrapolation in previous work on MPIs is limited in part by a network architecture that fixes the number of disparity planes during training and testing.
Increasing the renderable range of an MPI by simply increasing its fixed number of planes during training is not computationally feasible due to the memory limits of current GPUs.
We present a simple solution that increases disparity sampling frequency at test time by replacing the previously used 2D convolutional neural network (CNN) with a 3D CNN architecture and a randomized-resolution training procedure. We demonstrate that this change reduces the depth discretization artifacts found in distant views rendered by prior work, as shown in Figure~\ref{fig:teaser}.

\medskip
\noindent \textbf{Predicting disoccluded content for view extrapolation (Section~\ref{sec:disocclusions}).} We observe and explain why MPIs predicted by prior work~\cite{zhou18} contain approximately the same RGB content at each plane, and differ only in $\alpha$. This behavior results in unrealistic disocclusions with repeated texture artifacts, as illustrated in Figure~\ref{fig:teaser}. In general, the appearance of hidden scene content is inherently ambiguous, so training a network to simply minimize the distance between rendered and ground truth target views tends to result in unrealistic hallucinations of this occluded content. We propose to improve the realism of predicted disocclusions by constraining the appearance of occluded scene content at every depth to be drawn from visible scene points at or beyond that depth, and present a two-step MPI prediction procedure that enforces this constraint. We demonstrate that this strategy forces predicted disocclusions to contain plausible textures, alleviates the artifacts found in prior work, and produces more compelling extrapolated views than alternative approaches, as illustrated in Figures~\ref{fig:teaser} and~\ref{fig:results}. 

\section{Related Work}
\noindent \textbf{Traditional approaches for view synthesis.} View synthesis is an image-based rendering (IBR) task, with the goal of rendering novel views of scenes given only a set of sampled views. It is useful to organize view synthesis algorithms by the extent to which they use explicit scene geometry~\cite{shum00}. At one extreme are light field rendering~\cite{gortler96,levoy96} techniques, which require many densely sampled input images so that they can render new views by simply slicing the sampled light field without relying on accurate geometry estimation. At the other extreme are techniques such as view dependent texture mapping that rely entirely on an accurate estimated global mesh and then reproject and blend the texture from nearby input views to render new views~\cite{debevec96}.

Many successful modern approaches to view synthesis~\cite{chaurasia2013,hedman182,penner17,zitnick04} follow a strategy of computing detailed local geometry for each input view followed by forward projecting and blending the local texture from multiple input views to render a novel viewpoint. This research has traditionally focused on interpolating between input views and therefore does not attempt to predict content that is occluded in all input images. In contrast, we focus on the case of view extrapolation, where predicting hidden scene content is crucial for rendering compelling images.

\medskip
\noindent \textbf{Deep learning approaches for view synthesis.} Recently, a promising line of work has focused on training deep learning pipelines end-to-end to render novel views. One class of methods focuses on the challenging problem of training networks to learn about geometry and rendering from scratch and synthesize arbitrarily-distant views from such limited input as a single view~\cite{eslami18,park17,zhou16}. However, the lack of built-in geometry and rendering knowledge limits these methods to synthetic non-photorealistic scenarios. 

Other end-to-end approaches have focused on photorealistic view synthesis by learning to model local geometry from a target viewpoint and using this geometry to backwards warp and blend input views. This includes algorithms for interpolating between views along a 1D camera path~\cite{flynn16}, interpolating between four input corner views sampled on a plane~\cite{kalantari16}, and expanding a single image into a local light field of nearby views~\cite{srinivasan17}. These methods separately predict local geometry for every novel viewpoint and are not able to guarantee consistency between these predictions, resulting in temporal artifacts when rendering a sequence of novel views. Furthermore, the use of backward projection means that disoccluded regions must be filled in with replicas of visible pixels, so these techniques are limited in their ability to render convincing extrapolated views. 

The most relevant methods to our work are algorithms that predict a 3D scene representation from a source image viewpoint and render novel views by differentiably forward projecting this representation into each target viewpoint. This approach ensures consistency between rendered views and allows for the prediction of hidden content. Tulsiani \etal and Dhamo \etal predict a layered depth image (LDI) representation~\cite{dhamo18,tulsiani18}, but this approach is unable to approximate non-Lambertian reflectance effects. Furthermore, training networks to predict LDIs using view synthesis as supervision has proven to be difficult, and the training procedure requires a regularization term that encourages hidden content to resemble occluding content~\cite{tulsiani18}, limiting the quality of rendered disocclusions. Zhou \etal proposed the MPI scene representation~\cite{zhou18}, where novel views are rendered by forward projecting and alpha compositing MPI layers, and a deep learning pipeline is used to train an MPI prediction network using held-out views as supervision. They demonstrated that the MPI scene representation can convincingly render parallax and non-Lambertian effects for a small range of rendered views. We build upon this work and present a theoretical analysis of limits on views rendered from MPIs as well as a new MPI prediction framework that is able to render more compelling view extrapolations with disocclusions.

\medskip
\noindent \textbf{Inpainting occluded content.} Predicting the appearance of content hidden behind visible surfaces can be thought of as 3D scene inpainting. The problem of inpainting in 2D images has an extensive history~\cite{guillemot14}, ranging from early propagation techniques~\cite{bertalmio00} to modern CNN-based inpainting~\cite{yu18}. However, such algorithms must be applied separately to each rendering and therefore do not ensure consistency between different views of the same occluded content. 

A few recent works~\cite{baek16,howard14,philip18,thonat16} focus on multi-view inpainting, i.e.\ removing objects and inpainting the resulting empty pixels in a collection of multiple input images. This strategy operates on input image collections instead of scene representations, so it cannot be used to predict occluded content that only appears during view extrapolation. 

Finally, a recent line of work~\cite{firman16,song16,yang18} focuses on scene shape completion. These methods require an input depth image and only focus on inpainting the shape and semantics of hidden content and not its appearance, so the predicted scenes cannot be used for rendering novel views. In contrast to prior methods, our work addresses the problem of jointly inpainting the geometry, color, and opacity of hidden content in scenes to render convincing disocclusions.

\section{View Extrapolation for Visible Content}

\begin{figure}
\begin{center}
\newcommand{\width}{1.0\linewidth}
\includegraphics[width=\width]{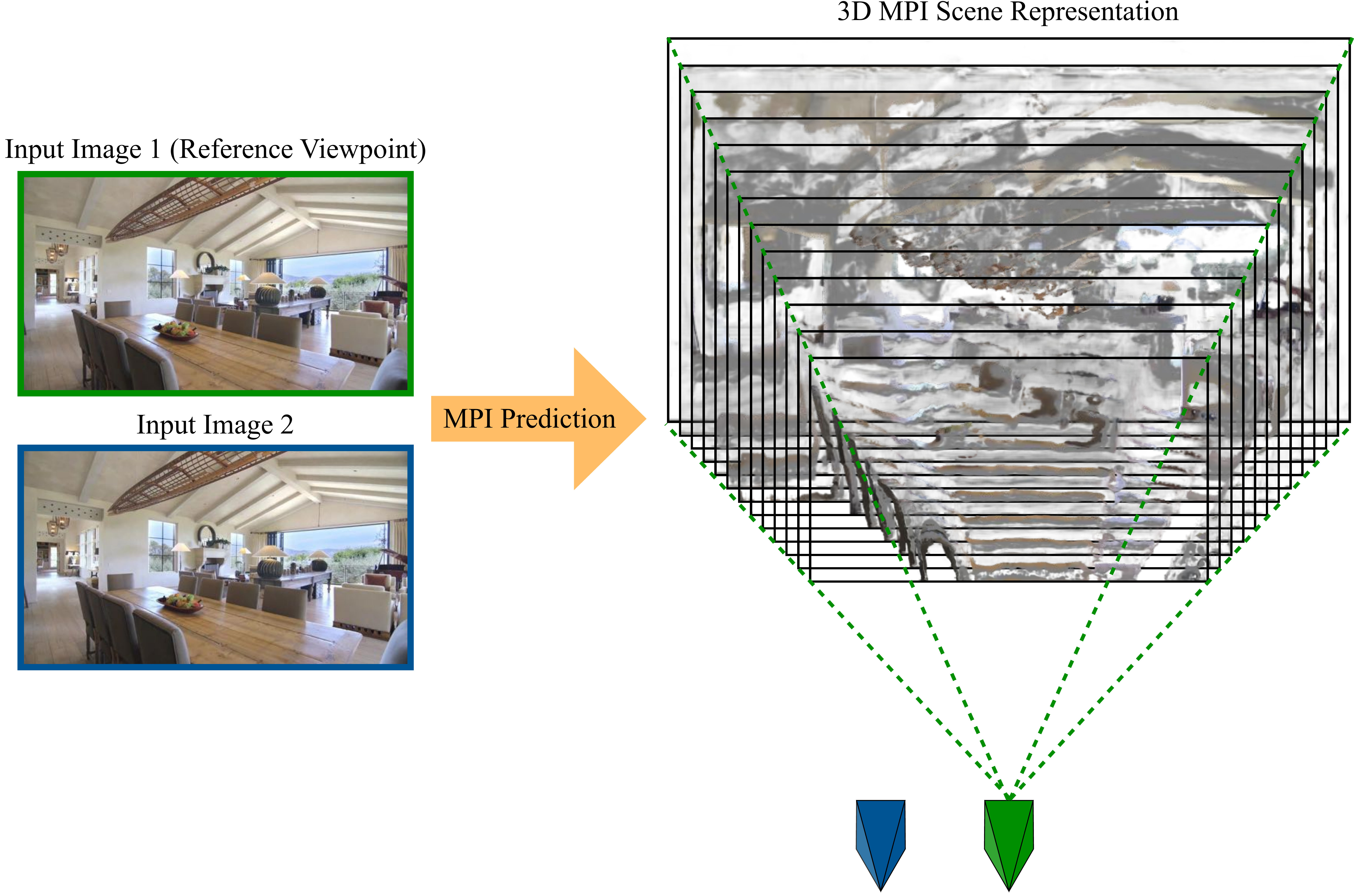} 
\caption{\textbf{MPI scene representation.} Our work builds on the MPI scene representation and prediction procedure introduced in~\cite{zhou18}. We train a deep network that takes two narrow-baseline images of a scene as input (captured at the blue and green camera poses shown above), and predicts an MPI scene representation, consisting of a set of fronto-parallel RGB$\alpha$ planes within a reference camera frustum (signified by the green camera above). Novel views are rendered by alpha compositing along rays from the MPI voxels into the novel viewpoint.}
\label{fig:representation}
\end{center}
\vspace{-0.15in}
\end{figure}

\subsection{MPI scene representation}

\begin{figure*}
\begin{center}
\newcommand{\width}{1.0\linewidth}
\includegraphics[width=\width]{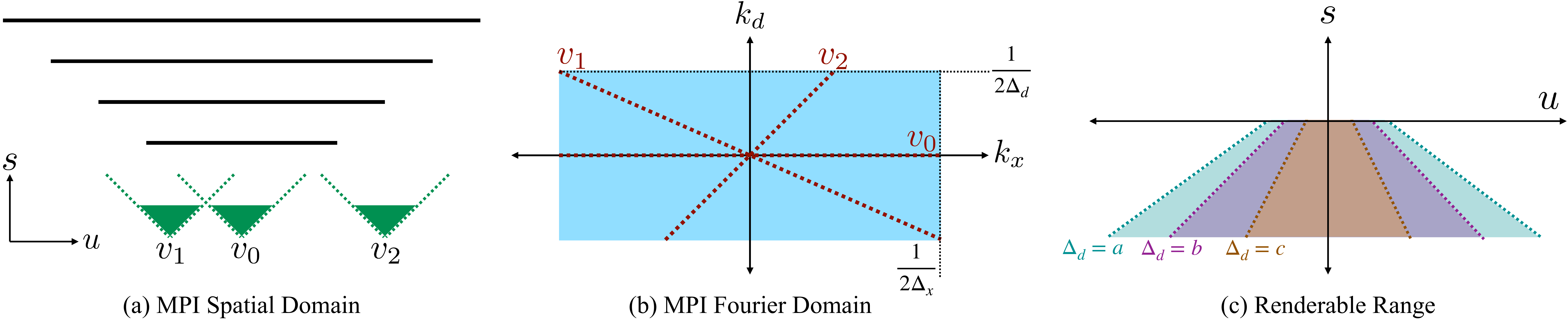} 
\caption{\textbf{Viewpoint limits for rendering visible content from an MPI.} Views rendered from an MPI without occlusions can be expressed as sheared integral projections of that MPI. (a) Here, we visualize a 2D slice from an MPI, where the $y$ dimension is constant and only the $x$ and $z$ dimensions vary. This MPI is in the reference viewpoint $v_0$. (b) In the frequency domain, rendered views are equivalent to 1D slices of the 2D MPI spectrum, where views further from the reference viewpoint correspond to Fourier slices at steeper slopes. The MPI spectrum is bandlimited due to its spatial and disparity sampling frequencies, so there is a range of viewpoints outside which rendered views will have a lower spatial bandwidth than the original MPI plane images. Viewpoint $v_1$ represents the maximum extent of this ``renderable range'', and $v_2$ represents a viewpoint outside this range. (c) The renderable range of views is shaped like a truncated cone, and we visualize how the range of renderable views shrinks linearly as we increase the disparity sampling interval $\Deltad$ from $a < b < c$.}
\label{fig:limits}
\end{center}
\vspace{-0.15in}
\end{figure*}

The multiplane image (MPI) scene representation, introduced by Zhou \etal~\cite{zhou18} and illustrated in Figure~\ref{fig:representation}, consists of a set of fronto-parallel RGB$\alpha$ planes within a reference camera's view frustum. An MPI can be thought of as a frustum-shaped volumetric scene representation where each ``voxel'' consists of a diffuse RGB color and opacity $\alpha$. Novel views are rendered from an MPI by alpha compositing the color along rays into the novel view using the ``over'' operator~\cite{levoy88,porter84}, which is easily implemented as homography-warping each MPI plane onto the sensor plane of the novel view (see Equation 2 in Zhou \etal~\cite{zhou18}), and alpha compositing the resulting images from back to front. 

\subsection{Theoretical signal processing limits for rendering visible content}
\label{sec:theory}

Perhaps surprisingly, there is a limit on views that can be rendered with high fidelity from an MPI, even if we just consider mutually-visible content, i.e., content visible from all input and target viewpoints. Rendering views beyond this limit results in depth discretization artifacts similar to aliasing artifacts seen in volume rendering~\cite{levoy88}.

We formalize this effect in the context of MPI renderings, and make use of Fourier theory to derive a bound on viewpoints that can be rendered from an MPI with high fidelity. Our model of rendering mutually-visible content from an MPI is conceptually similar to Frequency domain volume rendering~\cite{totsuka93} using a shear-warp factorization~\cite{lacroute94}. Additionally, our derivation of an MPI's ``renderable range'' is inspired by derivations for a 3D display's depth-of-field~\cite{zwicker06} and light field photography's ``refocusable range''~\cite{ng05}. Our main insight is that the 2D Fourier Transform of a view rendered from an MPI can be considered as a 2D slice through the 3D Fourier Transform of the MPI. An MPI is bandlimited by its fixed sampling frequency, so there exists a range of viewpoints outside of which rendered views will have a smaller spatial frequency bandwidth than the input images, potentially resulting in aliasing artifacts. We cover the main steps of this derivation below. Please refer to our supplementary materials for detailed intermediate steps and diagrams.

Let us consider rendering views from an MPI in the simplified case where (a) the camera is translated but not rotated, and (b) there is no occlusion, so all content is equally visible from every viewpoint. The rendered view $r_{\mathbf{u},s}(\mathbf{x})$ at a lateral translation $\mathbf{u}$ and axial translation $s$ relative to the reference camera center can then be expressed as:
\begin{equation}
\label{eq:render_visible}
\begin{split}
\resizebox{0.9\linewidth}{!}{$
\displaystyle
    r_{\mathbf{u},s}(\mathbf{x}) = \sum_{d \in \mathcal{D}} c(\mathbf{x'},d)
    = \sum_{d \in \mathcal{D}} c \left(\left(1-sd\right)\mathbf{x}+\mathbf{u}d,d \right)
    $}
\end{split}
\end{equation}
where $c(\mathbf{x},d)$ is the pre-multiplied RGB$\alpha$ at each pixel coordinate $\mathbf{x}$ and disparity plane $d$ within the set of MPI disparity planes $\mathcal{D}$. Note that $\mathbf{u}$ and $s$ are in units of pixels (such that the camera focal length $f=1$), and we limit $s$ to the range $-\infty<s<\sfrac{1}{\dmax}$ because renderings are not defined for viewpoints within the MPI volume. Additionally, note that the disparity $d$ is in units $\sfrac{1}{\mathit{pixel}}$.

To study the limits of views rendered from an MPI, let us consider a worst-case MPI with content in the subset of closest planes, for which we make a locally linear approximation to the coordinate transformation $(\mathbf{x},d) \to (\mathbf{x'},d)$:
\begin{equation}
\label{eq:render_approx}
    r_{\mathbf{u},s}(\mathbf{x}) = \sum_{d \in \mathcal{D}} c\left( \left(1-s \dmax\right)\mathbf{x}+\mathbf{u}d,d\right)
\end{equation}
where $\dmax$ is a constant. Now, we have expressed the rendering of mutually-visible content as a sheared and dilated integral projection of the MPI. We apply the generalized Fourier slice theorem~\cite{ng05} to interpret the Fourier transform of this integral projection as a 2D slice through the 3D MPI's Fourier transform. The resulting rendered view is the slice's inverse Fourier transform:
\begin{equation}
\label{eq:render_fslice}
r_{\mathbf{u},s}(\mathbf{x})=\mathcal{F}^{-1}\left\{C\left(\frac{k_{\mathbf{x}}}{1-s \dmax},\frac{-\mathbf{u}k_{\mathbf{x}}}{1-s \dmax}\right)\right\}
\end{equation}
where $\mathcal{F}^{-1}$ is the inverse Fourier transform and $C(k_{\mathbf{x}},k_{d})$ is the Fourier transform of $c(\mathbf{x},d)$.

An MPI is a discretized function, so the frequency support of $C$ lies within a box bounded by $\mypm\sfrac{1}{2\Deltabx}$ and $\mypm\sfrac{1}{2\Deltad}$, where $\Deltabx$ is the spatial sampling interval (set by the number of pixels in each RGB$\alpha$ MPI plane image) and $\Deltad$ is the disparity sampling interval (set by the number of MPI planes within the MPI disparity range). 

Figures~\ref{fig:limits}a and~\ref{fig:limits}b illustrate Fourier slices through the MPI's Fourier transform that correspond to rendered views from different lateral positions. Rendered views further from the reference view correspond to slices at steeper slopes. There is a range of slice slopes within which the spatial bandwidth of the rendered views is equal to that of the MPI, and outside of which the spatial bandwidth of the rendered views decreases linearly with the slice slope. 

We can solve for the worst-case ``renderable range'' by determining the set of slopes whose slices intersect the box in Figure~\ref{fig:limits}b at the spatial frequency boundary $\mypm\sfrac{1}{2\Deltabx}$. This provides constraints on camera positions $(\mathbf{u},s)$, within which rendered views enjoy the full image bandwidth:
\begin{equation}
s \leq 0,\quad \left|\mathbf{u}\right| \leq \frac{\Deltabx\left(1-s\dmax\right)}{\Deltad}
\end{equation}

Figure~\ref{fig:limits}c plots the renderable ranges with varying disparity intervals $\Deltad$, for an MPI with disparities up to $\dmax$. The allowed lateral motion extent increases linearly as the target viewpoint moves further axially from the MPI, starting at the reference viewpoint. Decreasing $\Deltad$ linearly increases the amount of allowed lateral camera movement. Intuitively, when rendering views at lateral translations from the reference viewpoint, the renderable range boundary corresponds to views in which adjacent MPI planes are shifted by a single pixel relative to each other before compositing.

\begin{figure*}
\begin{center}
\newcommand{\width}{1.0\linewidth}
\includegraphics[width=\width]{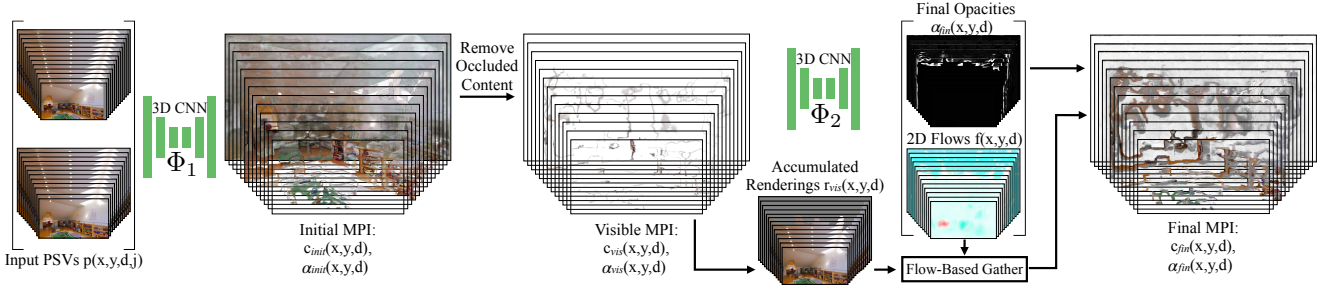} 
\caption{\textbf{Two-step MPI prediction pipeline.} We propose a two-step procedure to predict convincing hidden content in an MPI for view extrapolation. In the first step, a 3D CNN predicts an initial MPI from the input images' plane-sweep-volumes. Next, occluded content in this MPI is softly removed, resulting in a ``first-visible-surface'' MPI. In the second step, another 3D CNN predicts final MPI opacities and a 2D flow vector for each MPI voxel. The final MPI RGB colors are computed by using these predicted flows to gather RGB colors from back-to-front cumulative renderings of the visible content. This encourages hidden content at any depth to be synthesized by copying textures of visible content at or behind the same depth, which reduces the output space uncertainty for hidden content and thereby enables convincing view extrapolation with realistic disocclusions.}
\label{fig:pipeline}
\end{center}
\vspace{-0.15in}
\end{figure*}

\subsection{Increasing disparity sampling frequency with 3D CNN and randomized-resolution training}
\label{sec:arch}

Section~\ref{sec:theory} establishes that additional MPI planes increases the view extrapolation ability, and that this relationship is linear. Accordingly, the range of extrapolated views rendered by the original MPI method~\cite{zhou18} is limited because it uses a 2D CNN to predict a small fixed number of planes (32 planes at a spatial resolution of 1024$\times$576). Simply increasing this fixed number of planes in the network is computationally infeasible during training due to GPU memory constraints. Additionally, training on smaller spatial patches to allow for increased disparity sampling frequency prevents the network from utilizing larger spatial receptive fields, which is important for resolving depth in ambiguous untextured regions.

We propose a simple solution to predict MPIs at full resolution with up to 128 planes at test time by using a 3D CNN architecture, theoretically increasing the view extrapolation ability by $4\times$. The key idea is that because our network is fully 3D convolutional along the height, width, and depth planes dimensions, it can be trained on inputs with varying height, width, and number of depth planes. We use training examples across a spectrum of MPI spatial and disparity sampling frequencies that fit in GPU memory, ranging from MPIs with low spatial and high disparity sampling frequency (128 planes) to MPIs with high spatial and low disparity sampling frequency (32 planes). Perhaps surprisingly, we find that the trained network learns to utilize a receptive field equal to the maximum number of spatial and disparity samples it sees during training, even though no individual training example is of that size.

Our MPI prediction network takes as input a plane-sweep-volume tensor of size $[H,W,D,3N]$, where $H$ and $W$ are the image height and width, $D=|\mathcal{D}|$ is the number of disparity planes, and $N$ is the number of input images ($N=2$ in our experiments). This tensor is created by reprojecting each input image to disparity planes $\mathcal{D}$ in a reference view frustum. We use a 3D encoder-decoder network with skip connections and dilated convolutions~\cite{yu16} in the network bottleneck, so that the network's receptive field can encompass the maximum spatial and disparity sampling frequencies used during training. Please refer to our supplementary materials for a more detailed description of our network architecture and training procedure.

\section{View Extrapolation for Hidden Content}
\label{sec:disocclusions}

In the previous section, we described how view extrapolation is limited by the disparity sampling frequency, which is a fundamental property of the MPI scene representation. View extrapolation is also limited by the quality of hidden content, which is instead a property of the MPI prediction model. Models that train a CNN to directly predict an MPI from an input plane-sweep-volume (which contains homography-warped versions of the same RGB content at each plane) learn the undesirable behavior of predicting approximately the same RGB content at each MPI plane with variation only in $\alpha$ (see Figure 5 in Zhou \etal~\cite{zhou18}). We observe that this behavior is consistent for models that use either the original 2D CNN architecture or our 3D CNN architecture (Section~\ref{sec:arch}). Copies of the same RGB content at different MPI layers lead to ``repeated texture'' artifacts in extrapolated views, where disoccluded content contains repeated copies of the occluder, as visualized in Figure~\ref{fig:teaser}. 

We believe that this undesirable learned behavior is due to both the inductive bias of CNNs that directly predict an MPI from a plane-sweep-volume and the output uncertainty for disocclusions. The probability distribution over hidden scene content, conditioned on observed content, is highly multimodal---there may be many highly plausible versions of the hidden content. As a result, training a network to minimize the distance between rendered and ground truth views produces unrealistic predictions of disocclusions that are some mixture over the space of possible outputs.

We propose to reduce the output uncertainty by constraining the predicted hidden content at any depth, such that its appearance is limited to re-using visible scene content at or behind that depth. This effectively forces the network to predict occluded scene content by copying textures and colors from nearby visible background content. One possible limitation is that this constraint will have difficulty predicting the appearance of self-occlusions where an object extends backwards perpendicular to the viewing direction. However, as argued by the generic viewpoint assumption~\cite{freeman96}, it is unlikely that our reference viewpoint happens to view an object exactly at the angle at which it extends backwards along the viewing direction. In general, the majority of disoccluded pixels view background content instead of self-occlusions.

We enforce our constraint on the appearance of occluded content with a two-step MPI prediction procedure. The first step provides an initial estimate of the geometry and appearance of scene content visible from the reference viewpoint. The second step uses this to predict a final MPI where the color at each voxel is parameterized by a flow vector that points to a visible surface's color to copy.

In the first step, an input plane-sweep volume $p$ is constructed by reprojecting $j$ input images $i_{\mathbf{v}_j}$, each captured at a viewpoint $\mathbf{v}_j$, to disparity planes $d \in \mathcal{D}$. The 3D CNN $\neuralnet_1$ of Section~\ref{sec:arch} takes this plane-sweep volume and predicts an initial MPI's RGB and $\alpha$ values, $\cinit$ and $\ainit$:
\begin{equation}
\label{eq:initial_mpi}
\cinit(x,y,d),\ainit(x,y,d)=\neuralnet_{1}\big(p(x,y,d,j)\big).
\end{equation}
This initial MPI typically contains repeated foreground textures in occluded regions of the scene. In the second step of our procedure, we aim to preserve the predicted geometry and appearance of the first visible surface from the initial MPI while re-predicting the appearance and geometry of hidden content and enforcing our flow-based appearance constraint. We softly remove hidden RGB content from this initial MPI by multiplying each MPI RGB value by its transmittance $t$ relative to the reference viewpoint $\mathbf{v}_0$:
\begin{align}
t_{\mathbf{v}_0}(x,y,d)&= \ainit(x,y,d)\prod_{d\mathrlap{'}>d}\left[1-\ainit(x,y,d')\right] \label{eq:transmittance} \\
\cvis(x,y,d)&=\cinit(x,y,d)t_{\mathbf{v}_0}(x,y,d)\nonumber \\
\avis(x,y,d)&=t_{\mathbf{v}_0}(x,y,d)\label{eq:soft_removal}
\end{align}
where $\cvis$ and $\avis$ are the MPI RGB$\alpha$ planes from which content that is occluded from the reference view has been softly removed. Intuitively, a voxel's transmittance (Equation~\ref{eq:transmittance}) describes the extent to which an MPI voxel's color contributes to the rendered reference view. 

A second CNN $\neuralnet_2$ takes this reference-visible MPI, consisting of $\cvis$ and $\avis$, as input and predicts opacities $\afin(x,y,d)$ and a 2D flow vector for each MPI voxel $f(x,y,d)=[f_{x}(x,y,d),f_{y}(x,y,d)]$:
\begin{equation}
\resizebox{0.9\linewidth}{!}{$
\afin(x,y,d),f(x,y,d)=\neuralnet_2\big(\cvis(x,y,d),\avis(x,y,d)\big). \label{eq:final_prediction}
$}
\end{equation}
The final MPI's colors $\cfin(x,y,d)$ are computed by using these predicted flows to gather colors from renderings of the visible content at or behind each plane $\rvis(x,y,d)$:
\begin{align}
\rvis(x,y,d)&=\sum_{d' \leq d}\big[\cvis(x,y,d')\big] \label{eq:final_color_gather} \\
\cfin(x,y,d)&=\rvis\left(x+f_{x}(x,y,d),y+f_{y}(x,y,d),d\right). \nonumber
\end{align}
We gather the color from $\rvis$ using bilinear interpolation for differentiability. This constraint restricts the appearance of hidden content at each depth to be drawn from visible scene points at or beyond that depth.

\section{Training Loss}

As in Zhou \etal~\cite{zhou18}, we train our MPI prediction pipeline using view synthesis as supervision. Our training loss is simply the sum of reconstruction losses for rendering a held-out novel view $\rgt$ at target camera pose $\mathbf{v}_{t}$, using both our initial and final predicted MPIs. These MPIs are predicted from input images $i_{\mathbf{v}_0}$ and $i_{\mathbf{v}_1}$. We use a deep feature matching loss $\mathcal{L}_{\mathit{VGG}}$ for layers from the VGG-19 network~\cite{simoyan14}, using the implementation of Chen and Koltun~\cite{chen17}. The total loss $\mathcal{L}$ for each training example is:
\begin{equation}
\label{eq:training_loss}
\begin{split}
\mathcal{L}=&\mathcal{L}_{\mathit{VGG}}(\rinit(i_{\mathbf{v}_0},i_{\mathbf{v}_1},\mathbf{v}_{t}),\rgt)+\\&\mathcal{L}_{\mathit{VGG}}(\rfin(i_{\mathbf{v}_0},i_{\mathbf{v}_1},\mathbf{v}_{t}),\rgt)
\end{split}
\end{equation}
where $\rinit$ and $\rfin$ are rendered views from the initial and final predicted MPIs.

\section{Results}

The following section presents quantitative and qualitative evidence to validate the benefits of our method. Please view our supplementary video for rendered camera paths that demonstrate our predicted MPIs' ability to render high quality extrapolated views that are consistent with a 3D scene representation and contain realistic disocclusions.

\begin{table}
\begin{center}
\resizebox{\linewidth}{!}{
\Huge
\begin{tabular}{ l | ccc}
Algorithm & SSIM\textsubscript{fov} & SSIM\textsubscript{occ} & NAT\textsubscript{occ} \\
\hline
Original MPI~\cite{zhou18} & 0.838 & 0.803 & 0.805 \\
Our $\rinit$ &\cellcolor{Yellow}0.858 & 0.811 & 0.904 \\
$\rinit$ + Adversarial Disocclusion & 0.853 & 0.791 & 0.849 \\
Disocclusion Inpainting~\cite{yu18} & 0.808 & 0.691 & 0.227 \\
Our $\rfin$ & 0.853 &\cellcolor{Yellow}0.814 &\cellcolor{Yellow}0.931 \\
\end{tabular}
}
\vspace{1mm}
\caption{\textbf{Quantitative evaluation.} Images rendered from our predicted MPIs are quantitatively superior to those rendered from the original MPI model~\cite{zhou18}. Furthermore, our method predicts disocclusions that are both closer to the ground truth hidden content and more perceptually plausible than alternative methods.}
\label{table:quant}
\end{center}
\vspace{-0.1in}
\end{table}

\subsection{Experiment details} 

We train and evaluate on the open-source YouTube Real Estate 10K dataset~\cite{zhou18}\footnote{\url{https://google.github.io/realestate10k/}}, which contains approximately 10{,}000 YouTube videos of indoor and outdoor real estate scenes along with computed camera poses for each video frame. We generate training examples on the fly by sampling two source frames and a target frame from a randomly chosen video, so that the target image is not in between the source images (and therefore requires view extrapolation, not view interpolation) for $\sim$87\% of the training examples.

\begin{figure*}
\begin{center}
\newcommand{\width}{1.0\linewidth}
\includegraphics[width=\width]{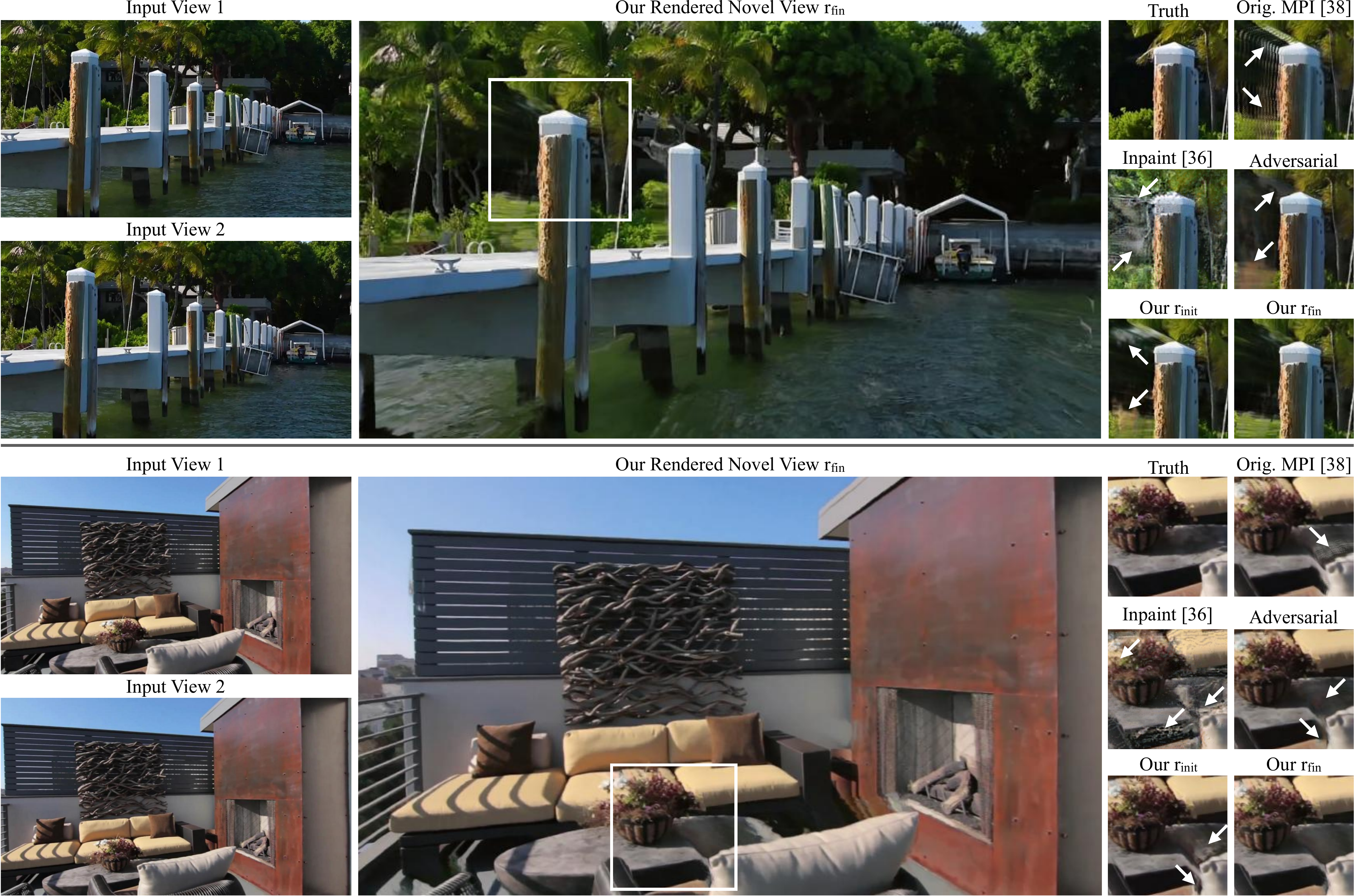} 
\caption{\textbf{Qualitative comparison of rendered novel views.} Our method predicts MPIs with convincing hidden content, as demonstrated by the disoccluded foliage textures to the left of the wooden pole in the top example, and the disoccluded region to the left of the grey pillow in the bottom example. Renderings from alternative methods contain depth discretization artifacts, implausible colors, blurry textures, and repeated textures in disoccluded regions.}
\label{fig:results}
\end{center}
\vspace{-0.1in}
\end{figure*}

The dataset is split into 9{,}000 videos for training and 1{,}000 for testing, where the test set videos do not overlap with those in the training dataset. From these test videos, we randomly sample 6{,}800 test triplets, each consisting of two input frames and a single target frame. 

\subsection{Evaluation metrics}

We use three metrics for our quantitative comparisons:

\noindent\textbf{SSIM\textsubscript{fov}:}\ To evaluate the overall quality of rendered images, we use the standard SSIM~\cite{wang04} metric computed over the region of the target image that views all MPI planes.

\noindent\textbf{SSIM\textsubscript{occ}:}\ To specifically assess the accuracy of predicted disocclusions, we evaluate SSIM over the subset of pixels that were not visible from the input reference viewpoint. We determine whether a pixel in a rendered target image is disoccluded by examining the MPI voxels that contribute to the rendered pixel's value, and thresholding the maximum change in transmittance of these contributing voxels between the reference and target viewpoint. Similarly to Equation~\ref{eq:transmittance}, we can compute the transmittance of each MPI voxel from a target viewpoint $\mathbf{v}_t$ as:
\begin{equation}
t_{\mathbf{v}_t}(x,y,d)= \alpha_{\mathbf{v}_t}(x,y,d)\prod_{d'>d}\left[1-\alpha_{\mathbf{v}_t}(x,y,d')\right]
\end{equation}
where $\alpha_{\mathbf{v}_t}$ is an MPI $\alpha$ plane homography-warped onto the sensor plane of viewpoint $\mathbf{v}_t$. We consider a pixel $(x,y)$ in the target rendered view as a member of the disoccluded pixels set $\mathcal{H}$ if the transmittance $t$ of any contributing MPI voxel is some threshold value greater than the same voxel's transmittance when rendering the reference viewpoint:
\begin{equation}
\resizebox{0.88\linewidth}{!}{$
\label{eq:disocc_map}
\displaystyle
\mathcal{H}=\left\{(x,y) : \max_{d} \big(t_{\mathbf{v}_t}(x,y,d)-t_{\mathbf{v}_0 \to \mathbf{v}_t}(x,y,d)\big) \geq \epsilon\right\}
$}
\end{equation}
where $t_{\mathbf{v}_0 \to \mathbf{v}_t}$ is the transmittance relative to the reference viewpoint, warped into the target viewpoint so that both transmittances are in the same reference frame. We compute disoccluded pixels using $\ainit$ for all models, to ensure that each model is evaluated on the same set of pixels. We set $\epsilon=0.075$ in our experiments. Please see our supplementary materials for visualizations of disoccluded pixels.

\noindent\textbf{NAT\textsubscript{occ}:}\ To quantify the perceptual plausibility of predicted disoccluded content, we evaluate a simple image prior over disoccluded pixels. We use the negative log of the Earth Mover's (Wasserstein-1) distance between gradient magnitude histograms of the rendered disoccluded pixels and the ground-truth pixels in each target image. Intuitively, realistic rendered image content should have a distribution of gradients that is similar to that of the true natural image~\cite{simoncelli97,weiss07}, and therefore a higher NAT\textsubscript{occ} score.

\subsection{Comparison to baseline MPI prediction}

We first show that renderings from both our initial and final predicted MPIs ($\rinit$ and $\rfin$) are superior to those from the original MPI method~\cite{zhou18}, which was demonstrated to significantly outperform other recent view synthesis methods~\cite{kalantari16,zhang15}. The increase in SSIM\textsubscript{fov} from ``Original MPI'' (Table~\ref{table:quant} row 2) to ``Our $\rinit$'' (row 3) demonstrates the improvement from our method's increased disparity sampling frequency. Furthermore, the increase in SSIM\textsubscript{occ} and NAT\textsubscript{occ} from ``Original MPI'' (row 2) to ``Our $\rfin$'' (row 6) demonstrates that our method predicts disoccluded content that is both closer to the ground truth and more plausible. Figure~\ref{fig:results} qualitatively demonstrates that renderings from our method contain fewer depth discretization artifacts than renderings from the original MPI work, and that renderings from our final MPI contain more realistic disocclusions without ``repeated texture'' artifacts.

\subsection{Evaluation of hidden content prediction}

We compare occluded content predicted by our model to the following alternative disocclusion prediction strategies:

\noindent\textbf{Our \textit{r\textsubscript{init}}:}\ We first compare renderings ``Our $\rfin$'' from our full method to the ablation ``Our $\rinit$'', which does not enforce our flow-based occluded content appearance constraint. The improvement in SSIM\textsubscript{occ} and NAT\textsubscript{occ} from Table~\ref{table:quant} row 3 to row 6 and the qualitative results in Figure~\ref{fig:results} demonstrate that our full method renders disocclusions that are both closer to the ground truth and more perceptually plausible with fewer ``repeated texture'' artifacts.

\noindent\textbf{``\textit{r\textsubscript{init}} + Adversarial Disocclusions'':}\ Next, we compare to an alternative two-step MPI prediction strategy. We use an identical $\neuralnet_1$ to predict the initial MPI, but $\neuralnet_2$ directly predicts RGB$\alpha$ planes instead of $\alpha$ and flow planes. We apply an adversarial loss to the resulting rendered target image to encourage realistic disocclusions (additional details in our supplementary materials). Table~\ref{table:quant} row 4 demonstrates that this strategy renders disocclusions that are less accurate but more perceptually plausible than the original MPI method, due to the adversarial loss. However, Figure~\ref{fig:results} demonstrates that the renderings from our full method contain sharper content and more accurate colors than those of the ``\textit{r\textsubscript{init}} + Adversarial Disocclusions'' strategy. We hypothesize that this is due to the difficulty of training a discriminator network when the number and location of ``fake'' disoccluded pixels varies drastically between training examples.

\noindent\textbf{``Disocclusion Inpainting'':}\ Finally, we compare to an image-based disocclusion prediction strategy. We remove the disoccluded pixels from our final MPI renderings and re-predict them using a state-of-the-art deep learning image inpainting model~\cite{yu18}. Table~\ref{table:quant} row 5 shows that this strategy results in an overall quality reduction, especially for the accuracy and plausibility of disoccluded regions. Figure~\ref{fig:results} visualizes the unrealistic inpainting results. Furthermore, as shown in our video, predicting disocclusions separately for each rendered image creates distracting temporal artifacts in rendered camera paths because the appearance of disoccluded content changes with the viewpoint. 

\section{Conclusion}

We have presented a theoretical signal processing analysis of limits for views that can be rendered from an MPI scene representation, and a practical deep learning method to predict MPIs that theoretically allow for $4\times$ more lateral movement in rendered views than prior work. This improvement is due to our method's ability to predict MPIs with increased disparity sampling frequency and our flow-based hidden content appearance constraint to predict MPIs that render convincing disocclusion effects. However, there is still a lot of room for improvement in predicting scene representations for photorealistic view synthesis that contain convincing occluded 3D content and are amenable to deep learning pipelines, and we hope that this work inspires future progress along this exciting research direction.

\small{\paragraph{Acknowledgments}
We thank John Flynn and Ben Mildenhall for fruitful discussions, and acknowledge support from ONR grant N000141712687, NSF grant 1617234, and an NSF fellowship. This work was done while PPS interned at Google Research.}

\clearpage
{\small
\bibliographystyle{ieee_fullname}
\bibliography{main}
}

\clearpage

\appendix
\setcounter{equation}{0}
\setcounter{figure}{0}
\setcounter{table}{0}
\setcounter{page}{1}
\makeatletter

\renewcommand{\theequation}{S\arabic{equation}}
\renewcommand{\thefigure}{S\arabic{figure}}
\renewcommand{\thetable}{S\arabic{figure}}
\renewcommand\@biblabel[1]{[S#1]}
\renewcommand\@cite[1]{[S#1]}

\section{Supplementary Video}

We have included a supplementary video that compares renderings from MPIs predicted by our model to those predicted by the original MPI method~\cite{ampi} and the ``Disocclusion Inpainting'' baseline that inpaints disoccluded pixels in each rendering using a state-of-the-art deep learning approach~\cite{ainpaint}. We invite readers to view this video for qualitative evidence of our method's ability to produce improved renderings with fewer depth discretization and repeated texture artifacts compared to the original MPI method, and with more convincing temporally-consistent disocclusions compared to the ``Disocclusion Inpainting'' baseline.

\section{Section 3.2 Derivation Details}

In this section, we provide additional details for the derivations in Section 3.2 of our main manuscript.

\begin{figure}
\begin{center}
\newcommand{\width}{1.0\linewidth}
\includegraphics[width=\width]{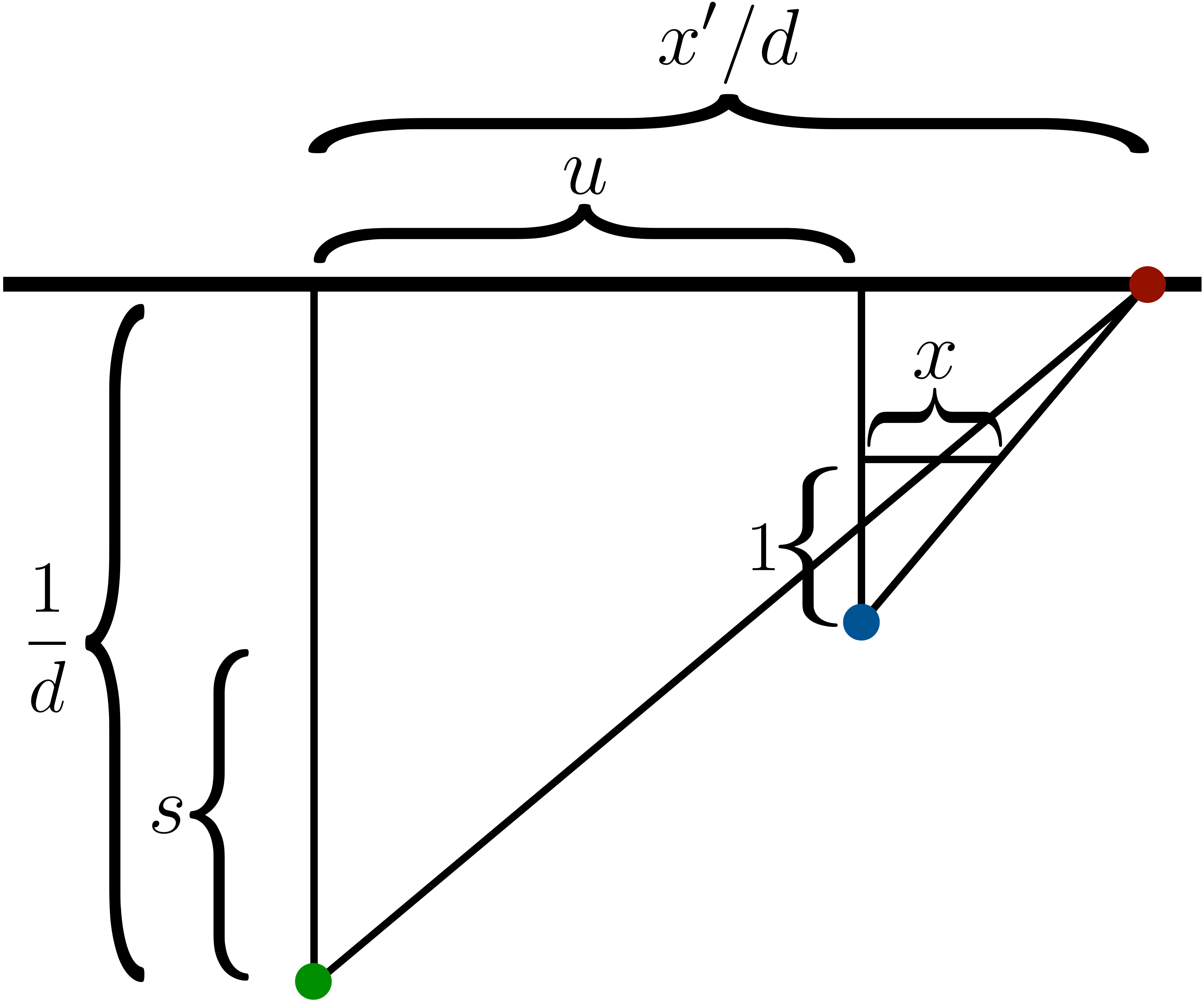} 
\caption{\textbf{Camera setup geometry.} An MPI in the frame of a reference camera (green dot) is viewed by a novel camera position (blue dot) at a translation $(u,s)$ relative to the reference camera.}
\label{fig:diagram}
\end{center}
\end{figure}

Figure~\ref{fig:diagram} illustrates a 2D slice of the camera setup geometry where we hold the $y$ dimension constant and view the $xz$-plane. An MPI in the frame of the reference camera (green dot) is viewed by a novel view camera (blue dot) at a translation $(u,s)$ relative to the reference camera. $x$ is the pixel coordinate of the red diffuse scene point on the blue camera's sensor plane, and $x'$ is the pixel coordinate of the red scene point on the visualized MPI plane at disparity $d$. Note that the MPI plane pixel coordinate $x'$ scales linearly with $\sfrac{1}{d}$, because each MPI plane RGB$\alpha$ image contains the same number of pixels sampled evenly within the camera frustum. It is straightforward to use the similar triangles of this diagram (above and to the right of the blue dot) to derive Equation 1 of our main manuscript:

\begin{equation}
\label{eq:render_visible}
\begin{split}
\resizebox{0.9\linewidth}{!}{$
\displaystyle
    r_{u,s}(x) = \sum_{d \in \mathcal{D}} c(x',d)
    = \sum_{d \in \mathcal{D}} c \left(\left(1-sd\right)x+ud,d \right)
    $}
\end{split}
\end{equation}
where $c(x,d)$ is the pre-multiplied RGB$\alpha$ at each pixel coordinate $x$ and disparity plane $d$ within the set of MPI disparity planes $\mathcal{D}$. Note that $u$ and $s$ are in units of pixels (such that the camera focal length $f=1$), and we limit $s$ to the range $-\infty<s<\sfrac{1}{\dmax}$ because renderings are not defined for viewpoints within the MPI volume. Additionally, note that the disparity $d$ is in units $\sfrac{1}{\mathit{pixel}}$.

We wish to study the limits of views rendered from an MPI, so let us consider a worst-case MPI with content in the subset of closest planes, for which we make a locally linear approximation to the coordinate transformation $(x,d) \to (x',d)$:
\begin{equation}
\label{eq:render_approx}
    r_{u,s}(x) = \sum_{d \in \mathcal{D}} c\left( \left(1-s \dmax\right)x+ud,d\right)
\end{equation}
where $\dmax$ is a constant. Now, we have expressed the rendering of mutually-visible content as a sheared and dilated integral projection of the MPI. We apply the generalized Fourier slice theorem~\cite{aslice} to interpret this integral projection of an MPI as a 2D slice through the 3D MPI's Fourier transform. Using operator notation, the generalized Fourier slice theorem can be expressed as:
\begin{equation}
\mathcal{F}^M \circ \mathcal{I}^N_M \circ \mathcal{B} \equiv \mathcal{S}^N_M \circ \frac{\mathcal{B}^{-\mathrm{T}}}{|\mathcal{B}^{-\mathrm{T}}|} \circ \mathcal{F}^N
\end{equation}
where $\mathcal{F}^M$ is the M-dimensional Fourier transform operator, $\mathcal{I}^N_M$ is the integral projection operator of an N-dimensional function to M dimensions by integrating out the last $N-M$ dimensions, $\mathcal{B}$ is a basis transformation operator (where $|\mathcal{B}^{-\mathrm{T}}|$ is the determinant of the inverse transpose of the transformation matrix), and $\mathcal{S}^N_M$ is the slicing operator that takes an M-dimensional slice from an N-dimensional function by setting the last $N-M$ dimensions to zero. The relevant values of the resulting transformation of MPI's Fourier transform, given the sheared and dilated MPI in Equation~\ref{eq:render_approx}, are:
\begin{equation}
    \begin{split}
    \mathcal{B}&=
        \begin{bmatrix}
        (1-sd_{max}) & u \\ 0 & 1
        \end{bmatrix} \\
    \mathcal{B}^{-\mathrm{T}}&=\frac{1}{1-sd_{max}}
        \begin{bmatrix}
        1 & 0 \\ -u & (1-sd_{max})
        \end{bmatrix}
    \end{split}
\end{equation}
We use these values to express the Fourier transformation of our sheared and dilated MPI as:
\begin{equation}
C(k_{x'},k_{d})=C\left(\frac{k_{x}}{1-s \dmax},\frac{-uk_{x}}{1-s \dmax}+k_d\right)
\end{equation}
where $C(k_{x},k_{d})$ is the Fourier transform of $c(\mathbf{x},d)$. We omit the $\sfrac{1}{|\mathcal{B}^{-\mathrm{T}}|}$ term since it is simply a scaling factor and can be absorbed into the definition of $C$. Finally, we compute the resulting rendered view as the inverse Fourier transform of the slice taken from the MPI's Fourier transformation by setting $k_d=0$:
\begin{equation}
\label{eq:render_fslice}
r_{u,s}(x)=\mathcal{F}^{-1}\left\{C\left(\frac{k_{x}}{1-s \dmax},\frac{-uk_{x}}{1-s \dmax}\right)\right\}
\end{equation}
where $\mathcal{F}^{-1}$ is the inverse Fourier transform. This connects our detailed supplementary derivation back with Equation 3 in the main manuscript.

\begin{figure}
\begin{center}
\newcommand{\width}{1.0\linewidth}
\includegraphics[width=\width]{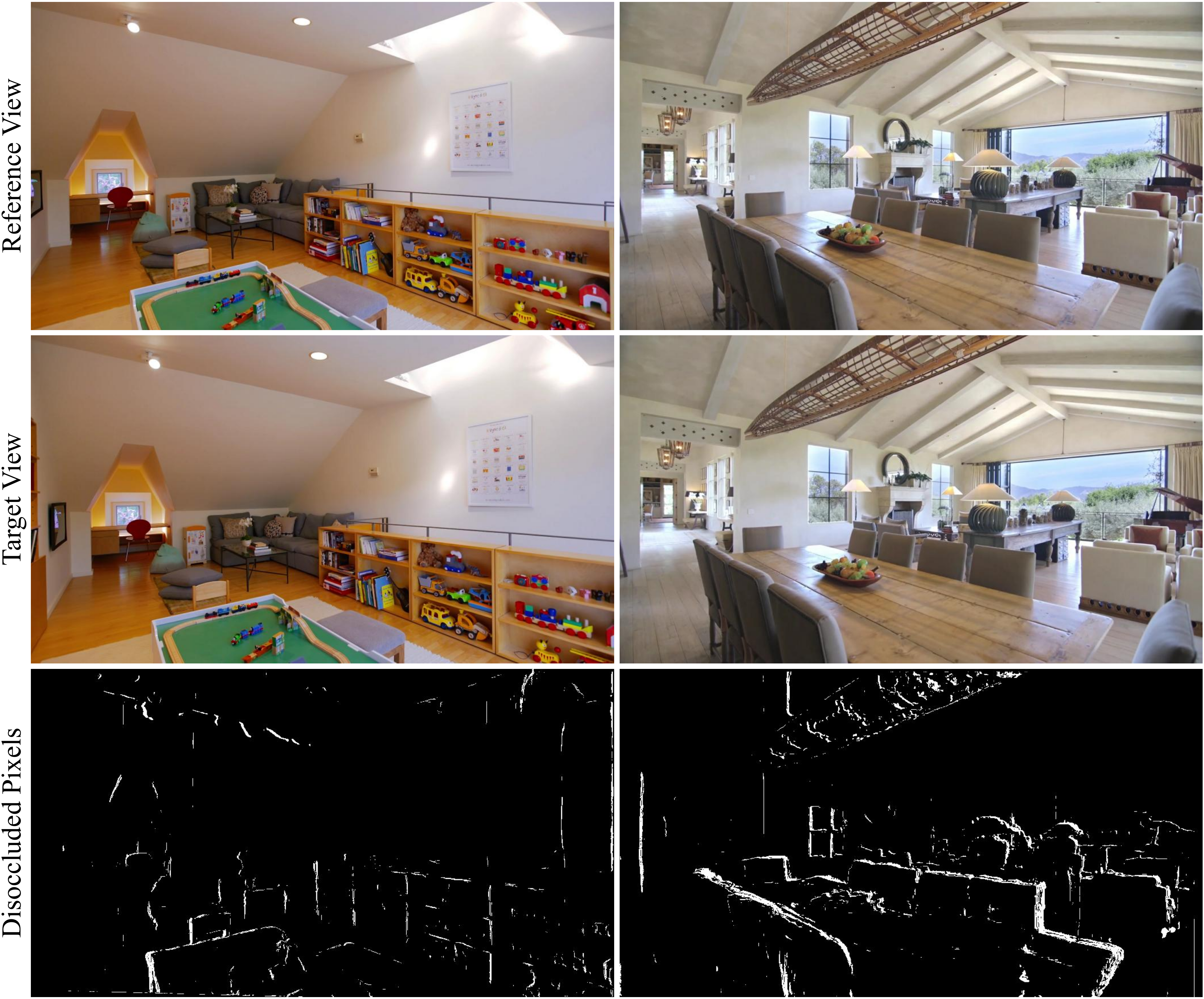} 
\caption{\textbf{Computed disocclusion masks.} We visualize example disocclusion masks for target view pixels that are occluded in the reference view, as used for our quantitative evaluations.}
\label{fig:disocclusion}
\end{center}
\vspace{-0.15in}
\end{figure}

\section{Network Architecture and Training Details}

Table~\ref{table:arch} contains precise specifications of the 3D convolutional neural network architecture described in Section 3.3 of the main manuscript.

We implement our system in TensorFlow~\cite{atensorflow}. We train using the Adam algorithm~\cite{aadam} for 300{,}000 iterations, with a learning rate of $2 \times 10^{-4}$, default parameters $\beta_1=0.9, \beta_2=0.999$, and a batch size of 1. 

For our randomized-resolution training, we uniformly sample input PSV tensors with sizes [height, width, \#planes, \#channels] of any of the following:

\begin{tabular}{@{\hspace{0.65in}}llll}
{[} 576,   &1024, & $2^4$, & 6{]} \\
{[} 576/2, &1024/2, & $2^5$, & 6{]} \\
{[} 576/4, &1024/4, & $2^5$, & 6{]} \\
{[} 576/4, &1024/4, & $2^6$, & 6{]} \\
{[} 576/4, &1024/4, & $2^7$, & 6{]} \\
{[} 576/8, &1024/8, & $2^5$, & 6{]} \\
{[} 576/8, &1024/8, & $2^6$, & 6{]} \\
{[} 576/8, &1024/8, & $2^7$, & 6{]} 
\end{tabular}

At test time, we apply this network on input PSV tensors with size [576,1024,$2^7$,6], which has the maximum spatial and depth resolutions seen during training.

\begin{table}[t]
\begin{center}
\resizebox{3.25in}{!}{

\begin{tabular}{ c | l | c c c c}

\hline \multicolumn{3}{c}{\textbf{Downsampling}}\\ \hline
 1-3 & ($3\times 3 \times 3$ conv, 8 features) $\times 3$ &
 $ H \times W \times D \times 8 $ \\ 
 
 4 & $3\times 3 \times 3$ conv, 16 features, stride 2 &
 $ H/2 \times W/2 \times D/2 \times 16 $ \\ 
 5-6 & ($3\times 3 \times 3$ conv, 16 features) $\times 2$  &
 $ H/2 \times W/2 \times D/2 \times 16 $ \\ 
 
 7 & $3\times 3 \times 3$ conv, 32 features, stride 2 &
 $ H/4 \times W/4 \times D/4 \times 32 $ \\ 
 8-9 & ($3\times 3 \times 3$ conv, 32 features) $\times 2$  &
 $ H/4 \times W/4 \times D/4 \times 32 $ \\ 
 
 10 & $3\times 3 \times 3$ conv, 64 features, stride 2 &
 $ H/8 \times W/8 \times D/8 \times 64 $ \\ 
 11-12 & ($3\times 3 \times 3$ conv, 64 features) $\times 2$  &
 $ H/8 \times W/8 \times D/8 \times 64 $ \\ 
 
 13 & $3\times 3 \times 3$ conv, 128 features, stride 2 &
 $ H/16 \times W/16 \times D/16 \times 128 $ \\ 
 14-15 & ($3\times 3 \times 3$ conv, 128 features) $\times 2$  &
 $ H/16 \times W/16 \times D/16 \times 128 $ \\ 
 
 \hline \multicolumn{3}{c}{\textbf{Bottleneck}}\\ \hline
 
 16 & $3\times 3 \times 3$ conv, 128 features, dilation rate 2 &
 $ H/16 \times W/16 \times D/16 \times 128 $ \\
 17 & $3\times 3 \times 3$ conv, 128 features, dilation rate 4 &
 $ H/16 \times W/16 \times D/16 \times 128 $ \\
 18 & $3\times 3 \times 3$ conv, 128 features, dilation rate 8 &
 $ H/16 \times W/16 \times D/16 \times 128 $ \\
 19 & $3\times 3 \times 3$ conv, 128 features &
 $ H/16 \times W/16 \times D/16 \times 128 $ \\
 
 \hline \multicolumn{3}{c}{\textbf{Upsampling}}\\ \hline
 
 20 & $2\times$ nearest neighbor upsample &
 $ H/8 \times W/8 \times D/8 \times 128 $ \\ 
 21 & concatenate 20 and 12 &
 $ H/8 \times W/8 \times D/8 \times (128+64) $ \\ 
 22-23 & ($3\times 3 \times 3$ conv, 64 features) $\times 2$ &
 $ H/8 \times W/8 \times D/8 \times 64 $ \\ 
 
 24 & $2\times$ nearest neighbor upsample &
 $ H/4 \times W/4 \times D/4 \times 64 $ \\ 
 25 & concatenate 24 and 9 &
 $ H/4 \times W/4 \times D/4 \times (64+32) $ \\ 
 26-27 & ($3\times 3 \times 3$ conv, 32 features) $\times 2$  &
 $ H/4 \times W/4 \times D/4 \times 32 $ \\ 
 
 28 & $2\times$ nearest neighbor upsample &
 $ H/2 \times W/2 \times D/2 \times 32 $ \\ 
 29 & concatenate 28 and 6 &
 $ H/2 \times W/2 \times D/2 \times (32+16) $ \\ 
 30-31 & ($3\times 3 \times 3$ conv, 16 features) $\times 2$  &
 $ H/2 \times W/2 \times D/2 \times 16 $ \\ 
 
 32 & $2\times$ nearest neighbor upsample &
 $ H \times W \times D \times 16 $ \\ 
 33 & concatenate 32 and 3 &
 $ H \times W \times D \times (16+8) $ \\ 
 34-35 & ($3\times 3 \times 3$ conv, 8 features) $\times 2$ &
 $ H \times W \times D \times 8 $ \\ 
 
 36 & $3\times 3 \times 3$ conv, 4 features (tanh) &
 $ H \times W \times D \times 4 $ \\ 

\end{tabular}
}
\vspace{0.02in}
\caption{\textbf{3D CNN network architecture.} Our initial MPI prediction CNN $\neuralnet_1$ uses the architecture described in the above table. Our final MPI prediction CNN $\neuralnet_2$ uses the same architecture without the bottleneck dilated convolutional layers. All convolutional layers in $\neuralnet_1$ and $\neuralnet_2$ use a ReLu activation, except for the final layer. $\neuralnet_1$ applies a tanh to all channels of the final layer, while $\neuralnet_2$ just applies a tanh on one output channel (corresponding to $\alpha$) and does not apply an activation to the predicted flows. 
\label{table:arch}}
\end{center}
\vspace{-0.15in}
\end{table}

\section{Disocclusion Mask Examples}

Figure~\ref{fig:disocclusion} visualizes disocclusion masks computed by our method (Equation 12 in the main manuscript) between pairs of reference and target viewpoints. 

\section{``\textit{r\textsubscript{init}} + Adversarial Disocclusions'' Details}

As described in our main manuscript, ``\textit{r\textsubscript{init}} + Adversarial Disocclusions'' is a two-step MPI prediction strategy we use as a baseline for comparisons. It uses an identical $\neuralnet_1$ to predict the initial MPI, but $\neuralnet_2$ directly predicts RGB$\alpha$ layers instead of flows, and we apply an adversarial loss to each final rendered target image to encourage realistic disocclusions. We adopt the SN-PatchGAN discriminator architecture with spectral normalization~\cite{aspectral} and a hinge loss objective function, as proposed in~\cite{ainpaint}. We add the adversarial loss to our main objective (Equation 10 in the main manuscript) with a weight $\lambda=5.0$.

\end{document}